\documentclass[letterpaper]{article} 
\usepackage{aaai2027}  
\nocopyright
\usepackage[hyphens]{url}  
\usepackage{graphicx} 
\usepackage{natbib}  
\usepackage{caption} 
\usepackage{amsmath}
\usepackage{amssymb}
\usepackage{booktabs}
\usepackage{multirow}
\usepackage{colortbl}
\usepackage{xspace}
\usepackage{makecell}
\usepackage{tabularx}

\newcommand{\proposed}{\textsc{G-STEER}\xspace}
\newcommand{\codeurl}{\url{https://github.com/SoojinY/G-STEER}\xspace}

\title{Personalized Deep Research Query Refinement\\with Graph-Scaffolded Evidence Grounding}
\author{
    Soojin Yoon\textsuperscript{\rm 1},
    Dongha Lee\textsuperscript{\rm 1}\corresponding
}
\affiliations{
    \textsuperscript{\rm 1}Department of Artificial Intelligence, Yonsei University\\
    \{soojiny,donalee\}@yonsei.ac.kr
}

\begin{document}

\maketitle

\begin{abstract}

User requests serve as research specifications for deep research agents, shaping what evidence to seek and how to synthesize it.
In personalized deep research, these specifications must additionally reflect user goals, constraints, preferences, and evaluation criteria.
User context can be incorporated either within the deep research pipeline or into the research specification provided as its input.
We focus on the latter, refining the user request into a personalized research specification before passing it to an unchanged deep research agent.
This requires resolving three coupled decisions: 
which framing factors are relevant, 
whether the available user context sufficiently supports them, 
and whether to retrieve user memory, ask the user, or stop and refine the query.
For training, G-STEER organizes framing factors as elicitation targets in an Intent Elicitation Graph that captures their dependencies.
It learns a clarification policy from graph-scaffolded trajectories spanning diverse factor dependencies and evidence conditions.
The policy produces a refined query while balancing target coverage against the costs of evidence acquisition.
Experiments show that G-STEER achieves the strongest overall weighted target coverage and the highest downstream report personalization across both evaluated DRAs, while asking roughly one third as many user questions as a strong clarification baseline.

\end{abstract}

\section{Introduction}
\label{sec:intro}
\begin{figure}[t]
    \centering
    \includegraphics[width=\linewidth]{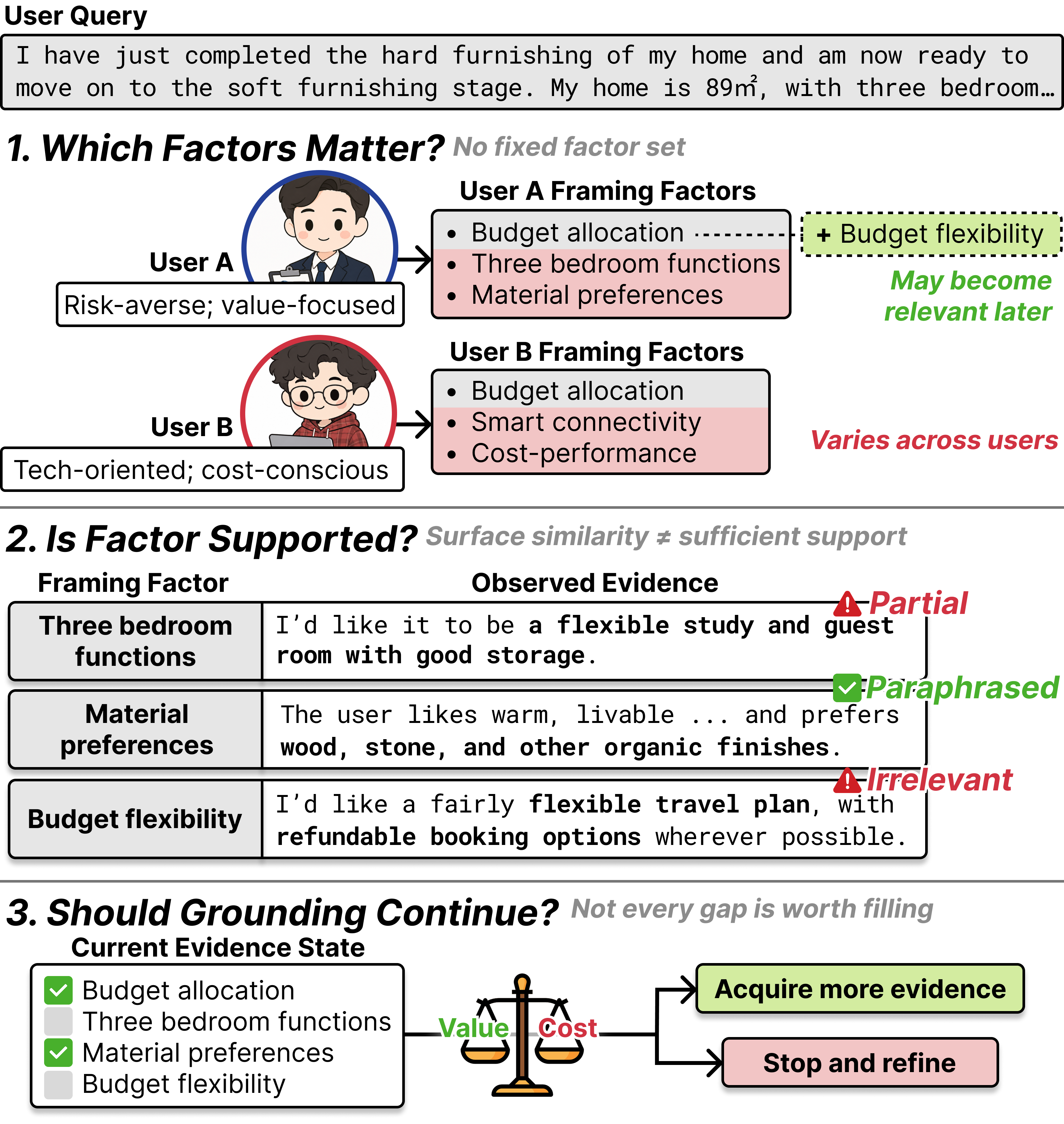}
    \caption{
    Three coupled decisions in personalized DR query refinement:
    which factors to include, whether they are supported, and whether to acquire evidence or stop.
    }
    \label{fig:motivation}
\end{figure}

Deep Research Agents (DRAs) have expanded information seeking from answering isolated queries to conducting multi-step investigations and generating comprehensive reports~\citep{du2025deepresearchbench,wang2026liveresearchbench}.
However, real-world research needs are rarely generic.
Even for the same request, users may expect different reports depending on their goals, constraints, preferences, and evaluation criteria.
This motivates Personalized Deep Research (PDR), where reports should reflect users' specific needs~\citep{pdrbench}.

Personalization can be introduced either within the DRA's internal pipeline~\citep{zhong2024memorybank,li2026personalized} or through its input research specification~\citep{cho2021personalized,zuo2023context}.
The former can directly shape retrieval and synthesis but requires access to internal components and system-specific adaptation, limiting its applicability to proprietary black-box DRAs.
We therefore study input-level refinement, where a refiner uses the initial query, user profile, persistent memory, and optional user interaction to produce a personalized research specification for an unchanged DRA.

Input-level refinement is related to pre-execution clarification, which elicits missing task information before downstream execution~\citep{qian2024tell,zhang2024askbeforeplan}.
IntentRL trains proactive clarification for open-ended deep research using graph-structured dialogue trajectories~\citep{intentrl}.
Whereas prior clarifiers primarily recover latent or missing task information through user interaction, PDR query refinement starts from a request that may already be sufficient for generic research and must determine which user-specific factors should alter its framing.
A relevant factor may already be supported by the user profile or persistent memory, require direct clarification, or not warrant further acquisition.
The refiner must therefore distinguish relevant from distracting context, grounded from weakly supported factors, and worthwhile from unnecessary interaction.

We formulate PDR query refinement as three coupled decisions: which user-specific factors should shape the final query (\emph{Framing Relevance}), whether the accumulated evidence sufficiently supports each factor (\emph{Evidential Sufficiency}), and whether to acquire additional evidence or stop and refine the query (\emph{Grounding Control}).
As illustrated in Figure~\ref{fig:motivation}, these decisions evolve jointly.
Resolving one factor may make another relevant, retrieved memory may remove the need for a question, and partial evidence may instead require direct clarification.
Learning these decisions requires interaction states in which factor relevance and evidential support change as evidence is acquired.

To this end, we propose \proposed, a training framework combining dependency-scaffolded trajectories with explicit evidence-state tracking.
\proposed organizes user-specific framing factors in an Intent Elicitation Graph (IEG), whose edges capture conditional elicitation dependencies.
Using the IEG and controlled evidence scenarios, \proposed constructs trajectories that vary conditional factor activation and evidence availability across the user profile, retrieved memory, and dialogue.
The resulting policy learns which factor to ground next, whether current evidence suffices, and whether to select \textsc{Retrieve}, \textsc{Ask}, or \textsc{Stop}.
The IEG is used only for trajectory construction and is unavailable at inference.

Supervised fine-tuning teaches evidence-state transitions, factor acquisition, and stopping behavior.
Target-anchored policy optimization then balances important factor coverage against retrieval and interaction costs.
At inference, the refiner induces and updates factor-specific evidence states from the initial query, user profile, retrieved memory, and dialogue without access to the offline elicitation targets or IEG.
When selecting \textsc{Stop}, it generates the personalized research specification from the accumulated evidence.

Experiments on PDR-Bench show that \proposed achieves the strongest overall weighted target coverage and the highest downstream report personalization across both evaluated DRAs.
It also asks roughly one third as many user questions as a strong clarification-and-rewrite baseline.

\begin{itemize}

    \item We formulate PDR query refinement as three coupled decisions and represent clarification dependencies among user-specific framing factors with an IEG.

    \item We train the refiner to choose \textsc{Retrieve}, \textsc{Ask}, or \textsc{Stop} based on the evidence available for each factor, using explicit evidence states and graph-scaffolded trajectories.

    \item We show that \proposed achieves the strongest downstream personalization across commercial black-box and open agent settings with substantially fewer questions than a strong clarification baseline.

\end{itemize}

\section{Related Works}
\label{sec:relwork}
\subsection{Personalizing Deep Research}

Personalized language models use user profiles or histories to adapt
outputs across classification, generation, and information-seeking tasks,
while memory-augmented agents retrieve past interactions to support
long-term dialogue~\citep{salemi2024lamp,salemi2024optimization,
zhong2024memorybank,packer2023memgpt,kim2026bespoke}.
Deep research extends personalization to long-horizon information seeking,
where user context can shape the research scope, evidence priorities, and
synthesis criteria.
PDR-Bench operationalizes this setting by pairing research tasks with
multiple user profiles, so that the same task is evaluated under different
user-specific perspectives~\citep{pdrbench}.
Its PQR framework measures personalization alignment (P), general content
quality (Q), and factual reliability (R): P assesses alignment with
user-specific needs, Q evaluates depth, coherence, and clarity, and R
measures factual support and citation coverage.
Recent PDR systems integrate user context into the DRA's internal pipeline,
including query development, retrieval, and report synthesis
~\citep{li2026personalized}.
We instead personalize the input research specification while leaving the
downstream DRA unchanged.

\subsection{Proactive Intent Clarification}

Prior work in information retrieval and conversational search resolves
underspecified requests through clarification questions, conversational
feedback, and iterative query rewriting
~\citep{zamani2020clarifying,cao2025icr}.
Recent work extends proactive clarification to language agents before
downstream execution.
Mistral-Interact detects vague instructions, asks users for missing intent,
and summarizes the clarified information into an actionable goal
~\citep{qian2024tell}.
CEP combines user clarification with environment interaction and tool use,
organizing missing details within constrained planning domains
~\citep{zhang2024askbeforeplan}.
IntentRL directly targets open-ended deep research, using an intent
refinement graph to construct clarification trajectories and train a
proactive asking policy~\citep{intentrl}.

These methods primarily recover information needed to make the expressed
task intent actionable.
Our setting instead personalizes the framing of a research request that may
already be sufficiently specified for generic execution.
Whereas CEP relies on domain-specific missing-detail structures and
IntentRL organizes clarification intents for proactive asking, the IEG
captures instance-specific dependencies among user-specific framing factors.
\proposed further distinguishes persistent memory from dialogue evidence and
jointly decides whether to \textsc{Retrieve}, \textsc{Ask}, or
\textsc{Stop} as evidence accumulates.

\section{Methodology}
\label{sec:method}
\subsection{Problem Formulation}

Given an initial user query $q$ and a static user profile $p_u$,
personalized query refinement aims to produce a refined query $q^\ast$
that preserves the original intent while making relevant user-specific
goals, constraints, preferences, and background explicit.
Because evidence for a framing factor may already be available in
$p_u$, recoverable from persistent user memory, or require direct
clarification, we formulate refinement as a sequential decision problem.
At step $t$, the agent conditions on $q$, $p_u$, and an accumulated
evidence state $\Sigma_t$, which records the grounding status, supported
value, and evidence source of each framing factor considered so far.

The agent selects
\[
a_t \in \{\textsc{Retrieve}, \textsc{Ask}, \textsc{Stop}\}.
\]
\textsc{Retrieve} searches persistent user memory,
\textsc{Ask} elicits evidence from the user, and
\textsc{Stop} terminates refinement and generates the refined query.
After \textsc{Retrieve} or \textsc{Ask}, the observation $o_{t+1}$
updates the evidence state:
\[
\Sigma_{t+1}
=
\operatorname{Update}(\Sigma_t, a_t, o_{t+1}).
\]
The policy is defined as
\[
\pi_\theta(a_t \mid q, p_u, \Sigma_t),
\]
and when it selects \textsc{Stop} at step $T$,
\[
q^\ast = g(q, p_u, \Sigma_T).
\]

\subsection{Framework Overview}
\label{sec:framework_overview}
\begin{figure*}[t]
    \centering
    \includegraphics[width=\textwidth]{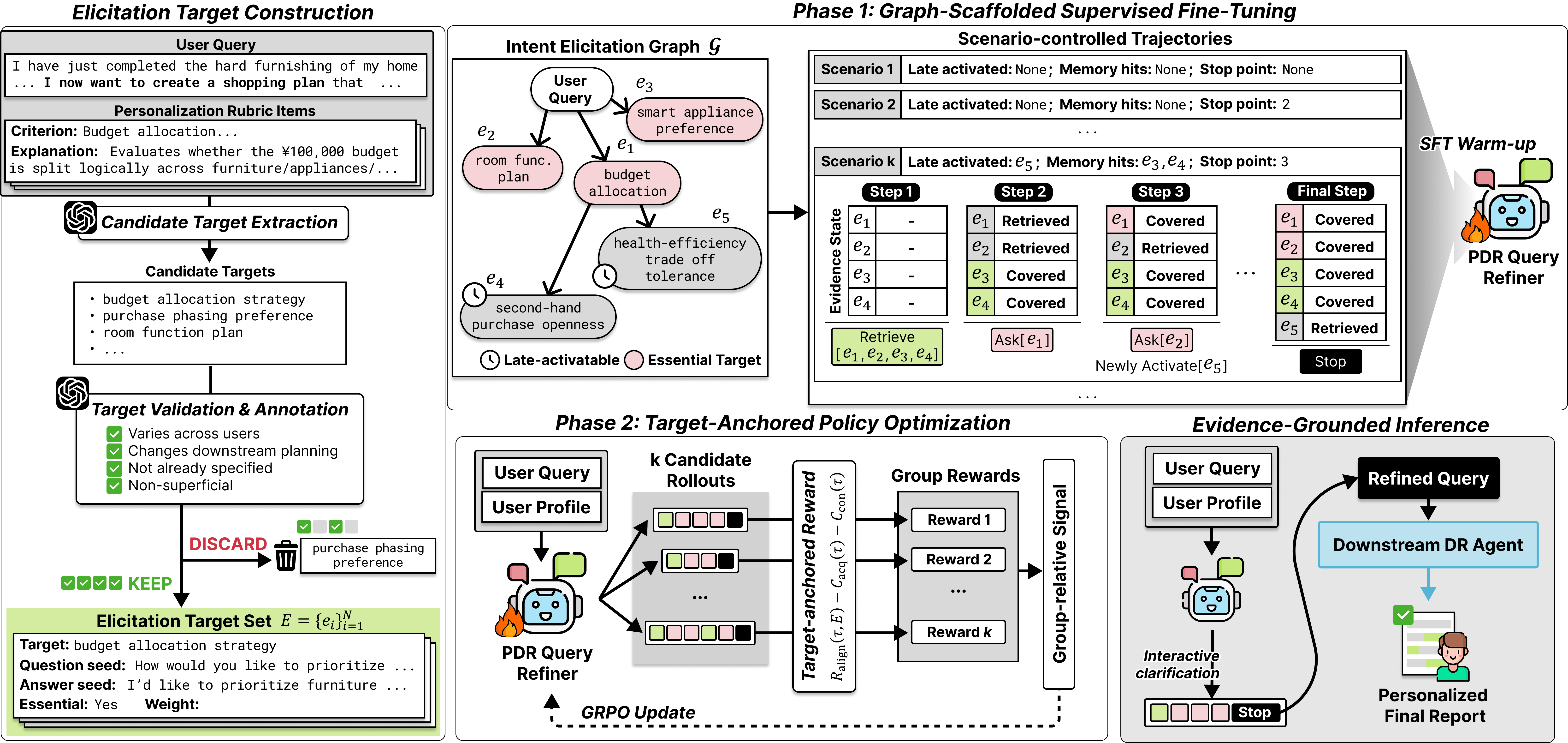}
    \caption{
    Overview of \proposed.
    }
    \label{fig:step_process}
\end{figure*}

\proposed is a training framework that constructs Graph-Scaffolded Trajectories for learning EvidencE-grounding decisions toward personalized deep research query Refinement.
\proposed trains a personalized query refiner in two stages.
First, it organizes user-specific elicitation targets into an
Intent Elicitation Graph (IEG) and constructs trajectories that vary target dependencies, evidence availability, and stopping conditions.
It then optimizes the clarification policy for target coverage and interaction efficiency.
At inference, the policy selects \textsc{Retrieve}, \textsc{Ask}, or \textsc{Stop} from the evolving evidence state without access to the elicitation targets or IEG.






\subsection{Elicitation Target Construction}

To supervise personalized query refinement, we represent user-dependent information that can alter downstream research planning as \emph{elicitation targets}.
For each instance, candidate elicitation targets are extracted from the initial query $q$ and its personalization rubric items. 
Rubric items are split only when they contain separable decisions that can independently change downstream planning, and the aspects that would naturally be resolved together remain merged.

The candidate targets are then validated.
A candidate is retained only when its answer can vary across users, materially affect downstream planning, and is not already specified by $q$.
Superficial, redundant, and overly narrow candidates are discarded.
Each retained target is also annotated as essential when leaving it unresolved would likely omit a major decision branch, apply inappropriate comparison criteria, or produce materially underspecified recommendations.

The resulting elicitation target set is $E=\{e_i\}_{i=1}^{N}$, where
$
e_i = (g_i, q_i^{\mathrm{seed}}, a_i^{\mathrm{seed}}, w_i, \eta_i).
$
Here, $g_i$ denotes the target description, 
$q_i^{\mathrm{seed}}$ and $a_i^{\mathrm{seed}}$ provide seed question--answer pair, 
$w_i$ is the rubric-derived importance weight, 
and $\eta_i$ is the essentiality label.
The weight $w_i$ is used only as a supporting signal because it is inherited from the benchmark evaluator.

\subsection{Graph-Scaffolded Supervised Fine-Tuning}
To train state-dependent clarification behavior, validated elicitation targets are organized into an \textit{Intent Elicitation Graph} (IEG) and instantiated as scenario-controlled trajectories.
These trajectories expose the policy to variations in target dependencies, evidence availability, and stopping conditions.

\paragraph{Intent Elicitation Graph.}
For each instance, an Intent Elicitation Graph is constructed as a directed acyclic graph $\mathcal{G}=(\mathcal{V},\mathcal{D})$, where $\mathcal{V}=E$ and the initial query acts as an external root.
An edge $(e_i,e_j)\in\mathcal{D}$ indicates that resolving $e_i$ can change whether or how $e_j$ should be elicited.
The graph therefore captures clarification dependencies rather than topical similarity.
A target is marked as late-activatable when it is initially inactive and becomes eligible only after one of its parents has been covered.

\paragraph{Scenario-Controlled Trajectories.}
Trajectory construction begins with a backbone subgraph containing all essential targets and their ancestors.
Additional trajectories include deterministic subsets of optional targets at ratios of $\{0, 0.5, 1.0\}$.
For each subgraph, targets are ordered so no child precedes its parent.
Multiple valid orders diversify the trajectories.
When several targets are eligible, optional targets closer to the essential backbone are prioritized.

Each IEG is instantiated into scenarios varying the target schedule, memory hits, late activation, and stopping point.
The trajectory first retrieves evidence for initially active targets.
After each memory or dialogue observation, the corresponding target slots are updated.
A late-activatable target remains inactive until an eligible parent slot is covered.
Once activated, memory is checked before asking the user.
Otherwise, the earliest uncovered target in the order is selected.
Each \textsc{Ask} action addresses one target at a time to reduce user burden.
The trajectory ends with \textsc{Stop} when all active slots are covered or the sampled stopping point is reached.

At step $t$, the evidence state $\Sigma_t$ consists of target slots recording coverage, values, and evidence from memory or dialogue.
\textsc{Plan} summarizes state changes and explains the next action.
\textsc{Execute} produces a retrieval query, clarification question, or refined query for \textsc{Retrieve}, \textsc{Ask}, or \textsc{Stop}, respectively.
At \textsc{Stop}, the teacher model refines the initial query using the final evidence state.

\paragraph{Policy Warm-up.}
The policy is fine-tuned on these trajectories to learn how to use the current evidence, choose between memory and user interaction, activate dependent targets, stop adaptively, and produce the required output for each action.
This initialization reduces the burden on reinforcement learning to discover both the interaction protocol and valid clarification behavior.

\subsection{Target-Anchored Policy Optimization}
\label{sec:grpo}
SFT teaches the clarification protocol, but does not directly optimize whether additional retrieval or user interaction is worthwhile.
We therefore refine the policy with trajectory-level Group Relative Policy Optimization (GRPO)~\citep{shao2024deepseekmath}.
Elicitation targets serve as reward anchors, allowing the policy to balance target coverage against interaction cost and protocol violations.

\paragraph{Trajectory-Level GRPO.}
For each training instance, the policy generates trajectories through interaction with a user simulator.
All trajectories start from the same memory and profile state.
Each trajectory $\tau$ receives the reward
\begin{align}
R(\tau,E)
=
R_{\mathrm{align}}(\tau,E)
-
C_{\mathrm{acq}}(\tau)
-
C_{\mathrm{con}}(\tau).
\end{align}
GRPO normalizes trajectory rewards within each group and applies the resulting advantage to all policy transitions in the corresponding rollout.
The update increases the likelihood of trajectories that cover important targets with fewer unnecessary actions and violations.

\paragraph{Target-Anchored Alignment.}
Let $V_T^{\mathrm{cov}}$ denote the covered slots in the final evidence state, and let $N$ be the number of elicitation targets.
Each elicitation target $e_i$ is represented by a reward anchor constructed from its target description $g_i$ and rubric criterion.
Each covered slot $v_j$ is represented by its name and purpose.

The similarity between target $e_i$ and slot $v_j$ is
$
S_{ij}=\max\left(0,\phi(g_i)^\top \phi(v_j)\right)
$
where $\phi(\cdot)$ is an L2-normalized encoder.
A one-to-one matching $\mathcal{M}^*$ maximizes the total similarity between targets and covered slots.
This ensures that each target and slot contributes to at most one match.

The alignment reward is
\begin{align}
R_{\mathrm{align}}(\tau,E) =
\alpha\,\mathrm{Cov}(\tau,E) - \beta\,C_{\mathrm{FP}}(\tau,E),
\end{align}
where
\begin{align}
\mathrm{Cov}(\tau,E)=
\frac{\sum_{(i,j)\in\mathcal{M}^{*}} w_i S_{ij}}
{\sum_{i=1}^{N} w_i}.
\end{align}
The coverage term gives greater reward to trajectories that cover important targets with strong alignment.

For each covered slot, let $\hat{s}_j=S_{ij}$ when $(i,j)\in \mathcal{M}^{*}$ and $\hat{s}_j=0$ otherwise.
The false-positive cost is
\begin{align}
C_{\mathrm{FP}}(\tau,E)
=
\frac{1}{N}
\sum_{v_j\in V_T^{\mathrm{cov}}}(1-\hat{s}_j).
\end{align}
It penalizes covered slots that are unmatched or weakly aligned with the elicitation targets.

\paragraph{Evidence Acquisition Cost.}
We penalize evidence acquisition to discourage unnecessary retrieval and user interaction.
The cost is
\begin{align}
C_{\mathrm{acq}}(\tau)
=
\lambda_{\mathrm{ask}}T_{\mathrm{ask}}
+
\lambda_{\mathrm{ret}}T_{\mathrm{ret}},
\end{align}
where $T_{\mathrm{ask}}$ and $T_{\mathrm{ret}}$ are the action counts, and $\lambda_{\mathrm{ask}}$ and $\lambda_{\mathrm{ret}}$ are their cost weights.
We set $\lambda_{\mathrm{ask}}>\lambda_{\mathrm{ret}}$ because asking the user imposes greater burden than retrieving from memory.

\paragraph{Behavioral Penalties.}
We penalize repeated actions and multi-target questions to reduce redundancy and preserve target-level atomicity.
The penalty is
\begin{align}
C_{\mathrm{con}}(\tau)
=
\lambda_{\mathrm{rep}}\,n_{\mathrm{rep}}(\tau)
+
\lambda_{\mathrm{multi}}\,n_{\mathrm{multi}}(\tau),
\end{align}
where $\lambda_{\mathrm{rep}}$ and $\lambda_{\mathrm{multi}}$ are the corresponding penalty weights.
$n_{\mathrm{rep}}(\tau)$ counts repeated acquisition attempts, including re-retrieving previously checked or covered slots, re-asking about previously queried slots, and generating questions whose embedding similarity to an earlier question exceeds a threshold.
$n_{\mathrm{multi}}(\tau)$ counts \textsc{Ask} actions that address multiple target slots or combine multiple question intents, identified from target slot annotations and lexical patterns.

\paragraph{Online User State Update.}

After each clarification trajectory, reusable evidence is selected from the final covered slots.
A slot is retained only when it is directly supported, specific to the user, useful for future clarification or personalization, and not already captured by the current profile.
Generic content, assistant advice, weak inferences, temporary details, and unnecessarily sensitive information are excluded.

Selected slots are stored as retrieval memories with their original names, purposes, and values.
The profile is updated with a one-sentence summary only when they support a durable, broadly reusable user pattern.
Otherwise, the profile remains unchanged.
The memories remain retrievable, while the profile serves as an optional summary grounded in them.

\subsection{Evidence-Grounded Inference}

At inference time, the trained policy refines the initial query \(q\) by iteratively acquiring user-specific evidence and deciding when to stop and generate the refined query \(q^*\).
The offline elicitation targets used for training and evaluation, as well as the IEG used to construct training trajectories, are not provided at inference.
The evidence state \(\Sigma_t\) defined above is represented at runtime as a textual set of slots for user-specific framing factors.
Each slot records the factor and its purpose, current grounding status, supported value, and evidence source, which may be the retrieved memory or dialogue.
Given \(q\) and the available user profile \(p_u\), the policy initializes \(\Sigma_0\) by inducing these slots through prompting rather than copying them from offline annotations.

Based on the current evidence state, the policy selects among \textsc{Retrieve}, \textsc{Ask}, and \textsc{Stop}.
After each \textsc{Retrieve} or \textsc{Ask} action, the resulting memory or dialogue observation may add, fill, or revise the relevant slots.
When \textsc{Stop} is selected at step \(T\), the policy generates \(q^*\) from \(q\), \(p_u\), and the final evidence state \(\Sigma_T\), with the prompt instructing it to incorporate only slot values supported by the available evidence.
\(q^*\) is passed to the downstream deep research agent.

\section{Experiments}
\label{sec:exp}
\begin{table}[t]
\centering
{
\small
\setlength{\tabcolsep}{3pt}
    \begin{tabular*}{\linewidth}{@{}l@{\extracolsep{\fill}}rrr@{}}
    \toprule
    \textbf{Split} & \textbf{Instances} & \makecell{\textbf{Total}\\\textbf{Targets}} & \makecell{\textbf{Avg. Targets}\\\textbf{per Instance}} \\
    \midrule
    \textbf{Train} & 105 (70\%) & 916 & 8.72 \\
    \textbf{Test} & 45 (30\%) & 418 & 9.29  \\
    \quad Seen user/task & 13 & 125 & 9.62  \\
    \quad User unseen & 12 & 119 & 9.92  \\
    \quad Task unseen & 10 & 82 & 8.20  \\
    \quad Both unseen & 10 & 92 & 9.20  \\
    \midrule
    \textbf{Total} & 150 & 1,334 & 8.89 \\
    \bottomrule
    \end{tabular*}
}
\caption{Dataset statistics.}
\label{tbl:data-statistics}
\end{table}

\begin{table*}[t]
\centering
{
\small
\setlength{\tabcolsep}{1mm}
\begin{tabular*}{\linewidth}{@{}l@{\extracolsep{\fill}}ccccc cc cc cc c@{}}
    \toprule
    \multirow{2.5}{*}{\textbf{Method}}
    & \multicolumn{5}{c}{\textbf{Overall}}
    & \multicolumn{2}{c}{\textbf{Unseen User}}
    & \multicolumn{2}{c}{\textbf{Unseen Task}}
    & \multicolumn{2}{c}{\textbf{Both Unseen}}
    & \multirow{2.5}{*}{\makecell{\textbf{Turns}\\Ask/Ret.}} \\
    \cmidrule(lr){2-6} \cmidrule(lr){7-8} \cmidrule(lr){9-10} \cmidrule(lr){11-12}
    & WCov & E-WCov & Prec. & Rec. & F1
      & WCov & F1
      & WCov & F1
      & WCov & F1
      & \\
    \midrule[\heavyrulewidth]

     Direct Rewrite
    & 0.3305 & 0.3697 & 0.3534 & 0.3134 & 0.3322
    & 0.3742 & 0.3624
    & 0.3654 & 0.3796
    & 0.3685 & 0.3730
    & 0.0/0.0 \\

     Memory-RAG Agent
    & 0.3180 & 0.3700 & 0.3968 & 0.3158 & 0.3517
    & 0.3451 & 0.3872
    & 0.4459 & \underline{0.4759}
    & 0.4801 & \underline{0.5025}
    & 0.0/1.5 \\

     Clarification Agent
    & 0.3399 & 0.3717 & 0.3754 & 0.3278 & 0.3499
    & 0.3415 & 0.3650
    & 0.3687 & 0.3947
    & 0.3704 & 0.3976
    & 3.5/0.0 \\

     Mem.+Clarif. Agent
    & 0.3822 & 0.4364 & \underline{0.4116} & 0.3565 & 0.3820
    & 0.3769 & 0.3715
    & 0.4505 & 0.4639
    & 0.4471 & 0.4294
    & 1.6/0.0 \\

    \midrule
    Mistral-Interact
    & 0.3811 & 0.4223 & 0.4099 & 0.3612 & \underline{0.3841}
    & 0.4152 & 0.3993
    & 0.4062 & 0.4416
    & 0.4771 & 0.4767
    & 3.6/0.0 \\

    CEP-Clarify+Rewrite
    & \underline{0.4060} & \underline{0.4497} & 0.3818 & \underline{0.3828} & 0.3823
    & 0.4026 & 0.3611
    & 0.4593 & \textbf{0.4889}
    & 0.4846 & 0.4698
    & 12.0/0.0 \\

    IntentRL-style Clarifier
    & 0.3760 & 0.4103 & 0.3983 & 0.3589 & 0.3775
    & \underline{0.4329} & \underline{0.4398}
    & \textbf{0.4825} & 0.4558
    & \underline{0.5242} & \textbf{0.5061}
    & 3.4/0.0 \\

    \midrule
    
     \proposed (Ours)
    & \textbf{0.4253} & \textbf{0.4639} & \textbf{0.4199} & \textbf{0.3971} & \textbf{0.4082}
    & \textbf{0.4886} & \textbf{0.4439}
    & \underline{0.4726} & 0.4359
    & \textbf{0.5266} & 0.4586
    & 4.1/1.2 \\
 
    \bottomrule
\end{tabular*}
}
\caption{Main evaluation results. Best scores are bolded and second-best scores are underlined. Avg. Turns reports the mean number of clarification-question and memory-retrieval turns (Ask/Ret.), excluding the final refinement turn.}
\label{tbl:main}
\end{table*}

\begin{table}[t]
\centering
{
\small
\setlength{\tabcolsep}{3pt}
\begin{tabular*}{\linewidth}{@{}l@{\extracolsep{\fill}}ccc@{}}
    \toprule
    \textbf{Method} & \textbf{WCov} & \textbf{E-WCov} & \textbf{F1} \\
    \midrule
    Full \proposed
    & \textbf{0.4253} & \underline{0.4639} & \underline{0.4082} \\
    \midrule
    \rowcolor{gray!15}
    \multicolumn{4}{l}{Untrained backbone comparison}\\

    \quad Qwen3-8B (Untrained)
    & 0.3761 & 0.4219 & 0.3892 \\

    \quad Qwen3-32B (Untrained)
    & 0.2966 & 0.3445 & 0.3316 \\

    \midrule
    \rowcolor{gray!15}
    \multicolumn{4}{l}{Training ablation}\\

    \quad Graph SFT only
    & \underline{0.4115} & 0.4412 & \textbf{0.4248} \\

    \quad GRPO only
    & 0.4029 & \textbf{0.4667} & 0.4036 \\

    \midrule

    \rowcolor{gray!15}
    \multicolumn{4}{l}{IEG Trajectory ablation}\\

    \quad Random-order SFT only
    & 0.3739 & 0.3729 & 0.3634 \\

    \quad Random-order SFT + GRPO
    & 0.3607 & 0.4124 & 0.3512 \\









    \bottomrule
\end{tabular*}
}
\caption{Ablation study comparing backbone scale, training stages, and IEG-scaffolded trajectory construction.}
\label{tbl:ablation}
\end{table}

\subsection{Experimental Setup}
\paragraph{Settings.}
We evaluate on the processed PDR-Bench~\citep{pdrbench}, split into training and test sets, as summarized in Table~\ref{tbl:data-statistics}.
Unless otherwise noted, we use Qwen3-8B~\citep{yang2025qwen3} as the backbone and fine-tune trainable variants with LoRA~\citep{hu2022lora}.
Qwen3-32B renders SFT trajectories and simulates users, while GPT-5.4~\citep{openai2026gpt54} constructs elicitation targets and IEGs.
We use \texttt{sentence-transformers/}\allowbreak\texttt{all-mpnet-base-v2}~\citep{reimers2019sentencebert,song2020mpnet} for semantic matching and FAISS-based user-memory indexing~\citep{douze2024faiss}.

At evaluation, memory and profiles are initialized from the final online states accumulated during GRPO on the training split and remain frozen across test instances.
Users absent from training start with empty states.
No held-out annotation or test interaction is used to construct or update these states.
Policies receive neither reference targets nor offline IEGs at inference.
See the appendix for details.


\paragraph{Evaluation.}
We evaluate each refined query against reference elicitation targets using precision, recall, and F1.
Because these metrics treat all targets equally, we also report weighted coverage (WCov), the normalized sum of PDR-Bench rubric-derived importance weights over covered targets, and E-WCov, its restriction to essential targets.
GPT-5.4 extracts user-specific framing factors from each refined query, and the extracted factors and reference targets are encoded with \texttt{sentence-transformers/}\allowbreak\texttt{all-mpnet-base-v2}~\citep{reimers2019sentencebert,song2020mpnet}.
A target is covered if its maximum cosine similarity to an extracted factor is at least 0.55.

Precision is the share of extracted factors matched to targets, recall the share of covered targets, and F1 their harmonic mean.
These intrinsic metrics assess target-level framing coverage and selectivity but do not directly verify whether every expressed factor value is entailed by the acquired evidence.
Using GPT-5.4 for both target construction and factor extraction may bias absolute scores, though all methods share the same evaluation pipeline.
Threshold sensitivity from 0.45 to 0.70 appears in the appendix.


\paragraph{Baselines.}
\textit{Generic prompting.}
Direct Rewrite rewrites the query without interaction.
We implement three agents with a shared ReAct-style interface~\citep{yao2023react}: Memory-RAG Agent uses \textsc{Retrieve}/\textsc{Stop}, Clarification Agent uses \textsc{Ask}/\textsc{Stop}, and Memory+Clarification Agent uses all three actions.
All prompting baselines use the same test-time profile as G-STEER, with no training or slot tracking.

\textit{Prior clarification.}
Mistral-Interact~\citep{qian2024tell} identifies underspecified instructions, elicits missing user intent, and returns the clarified goal as the final query.
CEP-Clarify+Rewrite~\citep{zhang2024askbeforeplan} adapts CEP's iterative clarification and uses Qwen3-8B to rewrite the dialogue history into the final query.
IntentRL-style Clarifier is a controlled adaptation of IntentRL~\citep{intentrl} that retains proactive clarification from dependency-structured trajectories.
To control for target-derived supervision, it uses the same IEG-scaffolded trajectories and interaction-cost objective as G-STEER, but is limited to \textsc{Ask}/\textsc{Stop} and refines the query from dialogue alone, without slot tracking.



\begin{table*}[t]
\centering
{
\small
\setlength{\tabcolsep}{1.5mm}
\begin{tabular}{@{}ll ccccc cccc@{}}
    \toprule
    \multirow{2.5}{*}{\makecell{\textbf{Downstream}\\\textbf{DRA}}}
    & \multirow{2.5}{*}{\textbf{Method}}
    & \multicolumn{5}{c}{\textbf{Personalization (P)}} 
    & \multicolumn{4}{c}{\textbf{Quality (Q)}} \\
    \cmidrule(lr){3-7} \cmidrule(lr){8-11}
    & & Overall & Goal & Content & Presentation & Actionability
    & Overall & Depth & Logic & Clarity \\
    \midrule
    \multirow{6}{*}{\makecell[l]{OpenAI\\DR}}
    & Original Query
    & 3.70 & 3.91 & 3.57 & 4.77 & 3.34
    & 5.43 & 5.13 & 5.72 & \underline{5.57}
    \\

    & Memory+Clarification
    & 3.81 & 4.10 & 3.92 & 4.73 & 3.32
    & 5.28 & 5.04 & 5.56 & 5.33
    \\

    & Mistral-Interact 
    & 3.28 & 3.52 & 3.28 & 4.46 & 2.84
    & 4.71 & 4.31 & 4.82 & 5.30
    \\

    & CEP-Clarify+Rewrite
    & 4.25 & \underline{4.55} & \underline{4.38} & 4.91 & 3.81
    & 5.40 & 5.15 & 5.62 & 5.54
    \\

    & IntentRL-style
    & 4.22 & 4.47 & 4.20 & \textbf{4.98} & \underline{3.85}
    & \textbf{5.69} & \textbf{5.53} & \textbf{5.84} & \textbf{5.80}
    \\

    \rowcolor{gray!12}
    \cellcolor{white} & \textbf{\proposed}
    & \textbf{4.35} & \textbf{4.65} & \textbf{4.46} & \underline{4.94} & \textbf{3.92}
    & \underline{5.56} & \underline{5.37} & \underline{5.80} & 5.54
    \\

    \midrule
    \multirow{6}{*}{OAgents}
    & Original Query
    & 5.05 & 4.98 & 4.32 & 5.96 & 5.22
    & \underline{6.73} & \underline{6.72} & \underline{7.13} & \underline{6.28}
    \\

    & Memory+Clarification
    & 5.12 & 5.14 & 4.68 & 5.94 & 5.12
    & 6.37 & 6.28 & 6.65 & 6.11
    \\

    & Mistral-Interact
    & 4.94 & 4.98 & 4.40 & 5.77 & 4.97
    & 6.05 & 5.94 & 6.43 & 5.84
    \\

    & CEP-Clarify+Rewrite
    & \underline{5.53} & \underline{5.58} & \underline{5.09} & \underline{6.04} & \underline{5.57}
    & 6.37 & 6.25 & 6.75 & 6.08
    \\

    & IntentRL-style
    & 4.93 & 4.91 & 4.45 & 5.51 & 5.03
    & 5.92 & 5.90 & 6.33 & 5.47
    \\

    \rowcolor{gray!12}
    \cellcolor{white} & \textbf{\proposed}
    & \textbf{5.70} & \textbf{5.72} & \textbf{5.28} & \textbf{6.31} & \textbf{5.73}
    & \textbf{6.77} & \textbf{6.74} & \textbf{7.22} & \textbf{6.31}
    \\

    \bottomrule
\end{tabular}
}
\caption{
    Downstream utility with fixed DR agents. P and Q measure personalization alignment and report quality, respectively.
}
\label{tbl:downstream-pq-full}
\end{table*}

\subsection{Main Results}
Table~\ref{tbl:main} shows that \proposed achieves the best overall performance across all coverage and matching metrics, with particularly strong results under user shift.
Under unseen-task evaluation, \proposed ranks second in WCov but trails CEP-Clarify+Rewrite in F1.
This indicates that its relative strength under task shift lies more in covering targets assigned greater importance than in uniform target matching.
On both-unseen instances, its relatively high WCov despite lower F1 suggests that the covered targets are concentrated among those with higher importance.
IntentRL-style uses the same rubric-derived supervision yet does not consistently outperform baselines without this supervision, suggesting that rubric exposure alone is insufficient.
Overall, \proposed achieves stronger performance than CEP-Clarify+Rewrite while asking 4.09 user questions on average, compared with 12.04 for CEP-Clarify+Rewrite, roughly one-third as many user questions.
The untrained Memory+Clarification baseline never invokes \textsc{Retrieve} despite access to both acquisition actions, showing an \textsc{Ask} bias rather than effective evidence-source routing.
Its difference from Clarification Agent instead reflects changes in the routing and stopping prompt, including a greater tendency to stop early, rather than any benefit from retrieved evidence.

\subsection{Ablation Studies}
\label{sec:ablation}
Table~\ref{tbl:ablation} examines backbone scale, training stages, and the role of IEG dependencies in trajectory construction.
The untrained Qwen3-32B model performs worse than Qwen3-8B across all metrics, showing that a larger backbone alone does not guarantee better selection of personalization factors.
Graph SFT achieves the highest F1, indicating that supervised trajectories stably teach which factors to acquire and include.
Relative to Graph SFT, the full model improves WCov by 0.0138 and E-WCov by 0.0227, while reducing F1 by 0.0166.
We interpret this as GRPO shifting the policy toward targets assigned greater importance rather than uniformly improving all matching metrics.
GRPO alone shows a related pattern, achieving the highest E-WCov but lower F1 than Graph SFT.
For the IEG ablation, we randomize target order while preserving the same target set, action conditions, and stopping scenarios.
SFT with randomized target order sharply reduces all metrics.
Adding GRPO improves E-WCov but further lowers WCov and F1.
Together, these results suggest that trajectory supervision following IEG dependencies substantially improves factor acquisition.

\begin{figure}[t]
    \centering
    \includegraphics[width=\linewidth]{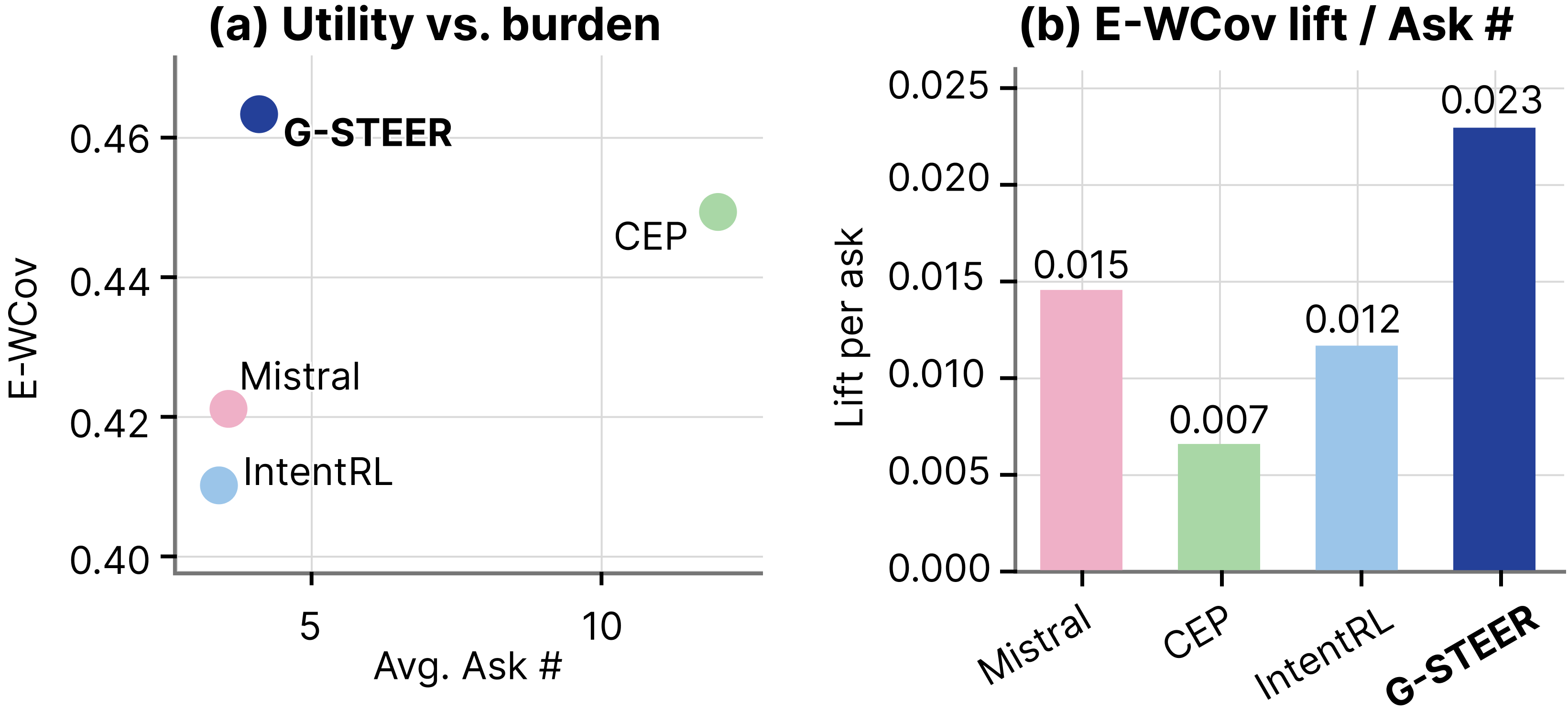}
    \caption{
    Utility--burden tradeoff among clarification methods.
    (a) E-WCov versus average questions;
    (b) E-WCov improvement over Direct Rewrite per question.
    }
    \label{fig:utility-burden}
\end{figure}

\subsection{Clarification Utility--Burden Trade-off}

Clarification methods can improve target coverage through useful evidence acquisition or simply by asking more questions, but each question imposes user effort.
Figure~\ref{fig:utility-burden} therefore compares E-WCov against the average number of questions and its improvement per question over Direct Rewrite.
We use E-WCov as the utility measure because it emphasizes factors essential to the personalized research specification.

G-STEER achieves the highest E-WCov and the largest improvement per question,
outperforming IntentRL-style at comparable burden and CEP-Clarify+Rewrite
with roughly one-third as many questions, indicating that its gains are
not merely due to asking more questions.

\begin{figure}[!t]
    \centering
    \includegraphics[width=0.919\linewidth]{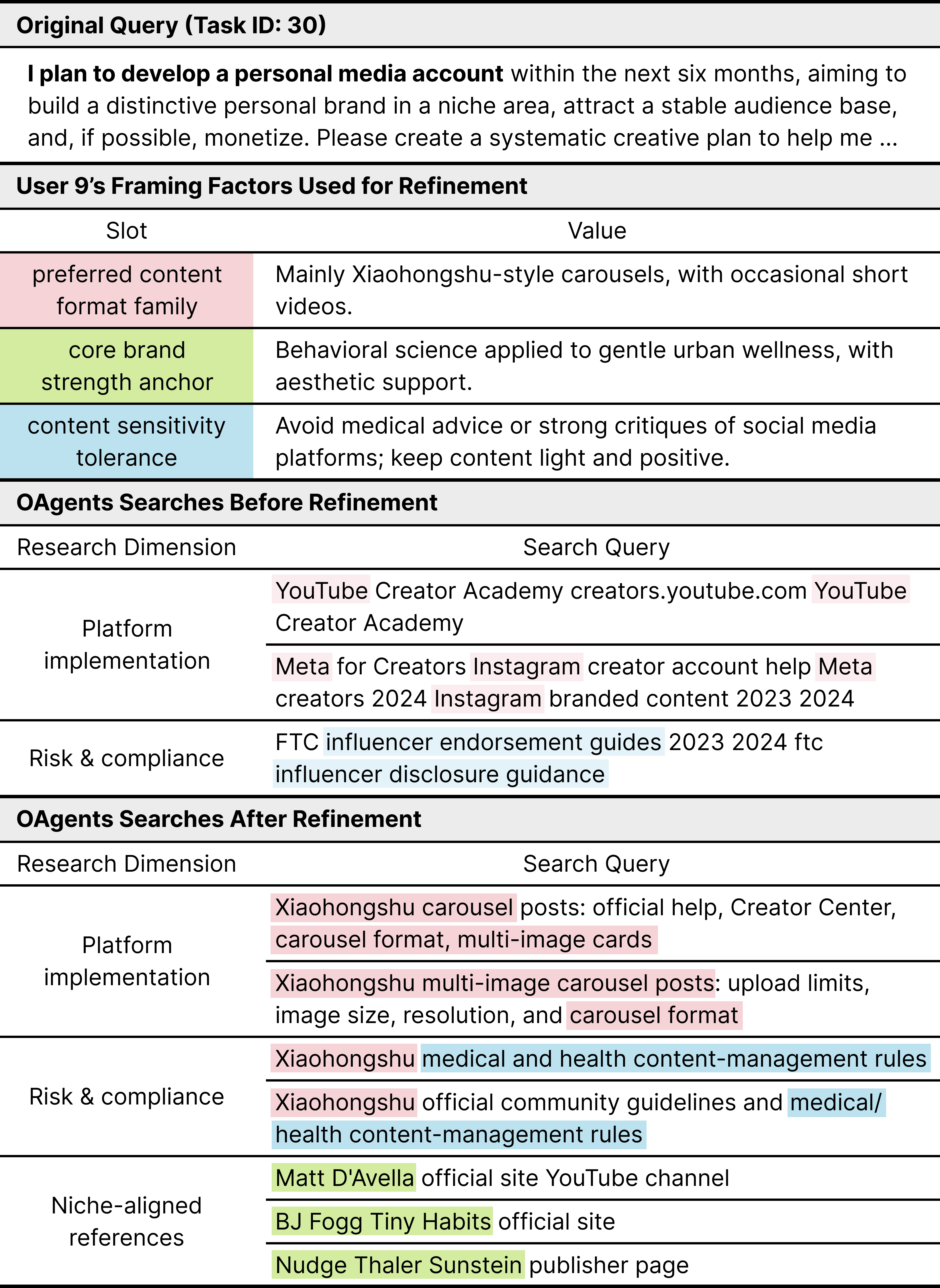}
    \caption{
    OAgents searches before and after refinement.
    Color families align factors to related search terms.
    }
    \label{fig:case_study}
\end{figure}

\subsection{Downstream Report Utility}

We complement intrinsic target matching with report-level evaluation under two fixed downstream settings: OpenAI Deep Research (\texttt{o4-mini-deep-research-}\allowbreak\texttt{2025-06-26}), a commercial black-box DRA, and OAgents, an open-source agent framework using \texttt{gpt-5-mini} for report generation~\citep{openai2026o4minideepresearch,zhu2025oagentsempiricalstudybuilding}.
Following PQR Evaluation~\citep{pdrbench}, we report personalization alignment (P) and content quality (Q), averaged over three evaluator rounds per report.
We omit factual reliability (R), which primarily reflects the fixed DRA's claim verification and citation behavior.
Because these report-level scores depend on both query quality and DRA-specific planning, retrieval, and synthesis, we compare methods within each DRA rather than absolute scores across DRAs.
Each DRA generates one report per method and test instance.

Table~\ref{tbl:downstream-pq-full} shows that \proposed achieves the highest overall P-score within both downstream settings.
With OpenAI Deep Research, \proposed outperforms CEP-Clarify+Rewrite in personalization while attaining the second-highest Q-score, showing that its personalization gains do not substantially compromise report quality.
With OAgents, it achieves the highest overall scores for both personalization and content quality.
These results show that \proposed consistently improves report personalization across distinct downstream systems without modifying their internals.

\subsection{Case Study}
Figure~\ref{fig:case_study} illustrates how factors specific to the user redirect the searches performed by OAgents.
Before refinement, OAgents searches broadly for guidance on creator platforms and general compliance requirements.
After refinement, it focuses on Xiaohongshu carousel formats and policies governing medical and health content, while also seeking references relevant to the user's intended positioning in gentle urban wellness.
These changes reflect the user's preferred format, content sensitivity, and brand direction.
The added reference branch shows that framing factors shape both search specificity and downstream research dimensions.

\section{Conclusion}
\label{sec:conclusion}

We introduced G-STEER, a framework for PDR query refinement that routes evidence acquisition between user memory and direct clarification.
It uses an Intent Elicitation Graph for dependency-consistent trajectory supervision and target-anchored optimization, improving factor acquisition and prioritizing important factors under interaction costs.
Across commercial black-box and open agent settings, \proposed achieves stronger report personalization with substantially fewer questions than the strongest clarification baseline.
Our evaluation relies on simulated users and a simple retrieval-based memory store, motivating future work with human interaction and richer memory systems.
Overall, effective personalized research requires identifying relevant context, grounding it appropriately, and knowing when to stop.



\bibliography{custom}

@inproceedings{du2025deepresearchbench,
      title={{DeepResearch Bench}: A Comprehensive Benchmark for Deep Research Agents},
      author={Mingxuan Du and Benfeng Xu and Chiwei Zhu and Licheng Zhang and Xiaorui Wang and Zhendong Mao},
      booktitle={International Conference on Learning Representations},
      year={2026},
      url={https://mlanthology.org/iclr/2026/du2026iclr-deepresearch/},
}

@inproceedings{wang2026liveresearchbench,
      title={{LiveResearchBench}: A Live Benchmark for User-Centric Deep Research in the Wild},
      author={Jiayu Wang and Yifei Ming and Riya Dulepet and Qinglin Chen and Austin Xu and Zixuan Ke and Frederic Sala and Aws Albarghouthi and Caiming Xiong and Shafiq Joty},
      booktitle={International Conference on Learning Representations},
      year={2026},
      url={https://mlanthology.org/iclr/2026/wang2026iclr-liveresearchbench/},
}

@misc{intentrl,
      title={IntentRL: Training Proactive User-intent Agents for Open-ended Deep Research via Reinforcement Learning}, 
      author={Haohao Luo and Zexi Li and Yuexiang Xie and Wenhao Zhang and Yaliang Li and Ying Shen},
      year={2026},
      eprint={2602.03468},
      archivePrefix={arXiv},
      primaryClass={cs.AI},
      url={https://arxiv.org/abs/2602.03468}, 
}

@inproceedings{pdrbench,
      title={Towards Personalized Deep Research: Benchmarks and Evaluations}, 
      author={Yuan Liang and Jiaxian Li and Yuqing Wang and Piaohong Wang and Motong Tian and Pai Liu and Shuofei Qiao and Runnan Fang and He Zhu and Ge Zhang and Minghao Liu and Yuchen Eleanor Jiang and Ningyu Zhang and Wangchunshu Zhou},
      booktitle={International Conference on Learning Representations},
      year={2026},
      url={https://mlanthology.org/iclr/2026/liang2026iclr-personalized/}, 
}

@inproceedings{li2026personalized,
author = {Li, Xiaopeng and Zhang, Wenlin and Zhang, Yingyi and Jia, Pengyue and Wang, Yejing and Wang, Yichao and Liu, Yong and Guo, Huifeng and Zhao, Xiangyu},
title = {Personalized Deep Research: A User-Centric Framework, Dataset, and Hybrid Evaluation for Knowledge Discovery},
year = {2026},
isbn = {9798400725999},
publisher = {Association for Computing Machinery},
address = {New York, NY, USA},
url = {https://doi.org/10.1145/3805712.3808609},
doi = {10.1145/3805712.3808609},
booktitle = {Proceedings of the 49th International ACM SIGIR Conference on Research and Development in Information Retrieval},
pages = {3284–3291},
numpages = {8},
location = {Australia},
series = {SIGIR '26}
}

@inproceedings{cho2021personalized,
    title = "Personalized Search-based Query Rewrite System for Conversational {AI}",
    author = "Cho, Eunah  and
      Jiang, Ziyan  and
      Hao, Jie  and
      Chen, Zheng  and
      Gupta, Saurabh  and
      Fan, Xing  and
      Guo, Chenlei",
    editor = "Papangelis, Alexandros  and
      Budzianowski, Pawe{\l}  and
      Liu, Bing  and
      Nouri, Elnaz  and
      Rastogi, Abhinav  and
      Chen, Yun-Nung",
    booktitle = "Proceedings of the 3rd Workshop on Natural Language Processing for Conversational AI",
    month = nov,
    year = "2021",
    address = "Online",
    publisher = "Association for Computational Linguistics",
    url = "https://aclanthology.org/2021.nlp4convai-1.17/",
    doi = "10.18653/v1/2021.nlp4convai-1.17",
    pages = "179--188"
}

@inproceedings{zuo2023context,
    title = "Context-Aware Query Rewriting for Improving Users' Search Experience on {E}-commerce Websites",
    author = "Zuo, Simiao  and
      Yin, Qingyu  and
      Jiang, Haoming  and
      Xi, Shaohui  and
      Yin, Bing  and
      Zhang, Chao  and
      Zhao, Tuo",
    editor = "Sitaram, Sunayana  and
      Beigman Klebanov, Beata  and
      Williams, Jason D",
    booktitle = "Proceedings of the 61st Annual Meeting of the Association for Computational Linguistics (Volume 5: Industry Track)",
    month = jul,
    year = "2023",
    address = "Toronto, Canada",
    publisher = "Association for Computational Linguistics",
    url = "https://aclanthology.org/2023.acl-industry.59/",
    doi = "10.18653/v1/2023.acl-industry.59",
    pages = "616--628"
}

@inproceedings{kim2026bespoke,
      title={BESPOKE: Benchmark for Search-Augmented Large Language Model Personalization via Diagnostic Feedback}, 
      author={Hyunseo Kim and Sangam Lee and Kwangwook Seo and Dongha Lee},
      booktitle={International Conference on Machine Learning},
      year={2026},
      url={https://icml.cc/virtual/2026/poster/64103}, 
}

@inproceedings{salemi2024lamp,
  title = {{LaMP}: When Large Language Models Meet Personalization},
  author = {Salemi, Alireza and Mysore, Sheshera and Bendersky, Michael and Zamani, Hamed},
  booktitle = {Proceedings of the 62nd Annual Meeting of the Association for Computational Linguistics (Volume 1: Long Papers)},
  pages = {7370--7392},
  year = {2024},
  publisher = {Association for Computational Linguistics},
  url = {https://aclanthology.org/2024.acl-long.399/},
  doi = {10.18653/v1/2024.acl-long.399}
}

@inproceedings{salemi2024optimization,
  title = {Optimization Methods for Personalizing Large Language Models through Retrieval Augmentation},
  author = {Salemi, Alireza and Kallumadi, Surya and Zamani, Hamed},
  booktitle = {Proceedings of the 47th International ACM SIGIR Conference on Research and Development in Information Retrieval},
  pages = {752--762},
  year = {2024},
  publisher = {ACM},
  url = {https://doi.org/10.1145/3626772.3657783},
  doi = {10.1145/3626772.3657783}
}

@inproceedings{zhong2024memorybank,
  title = {{MemoryBank}: Enhancing Large Language Models with Long-Term Memory},
  author = {Zhong, Wanjun and Guo, Lianghong and Gao, Qiqi and Ye, He and Wang, Yanlin},
  booktitle = {Proceedings of the AAAI Conference on Artificial Intelligence},
  volume = {38},
  pages = {19724--19731},
  year = {2024},
  doi = {10.1609/aaai.v38i17.29946},
  url = {https://ojs.aaai.org/index.php/AAAI/article/view/29946}
}

@misc{packer2023memgpt,
  title = {{MemGPT}: Towards {LLM}s as Operating Systems},
  author = {Packer, Charles and Wooders, Sarah and Lin, Kevin and Fang, Vivian and Patil, Shishir G. and Stoica, Ion and Gonzalez, Joseph E.},
  year = {2023},
  eprint = {2310.08560},
  archivePrefix = {arXiv},
  primaryClass = {cs.AI},
  url = {https://arxiv.org/abs/2310.08560}
}

@inproceedings{yao2023react,
  title = {{ReAct}: Synergizing Reasoning and Acting in Language Models},
  author = {Yao, Shunyu and Zhao, Jeffrey and Yu, Dian and Du, Nan and Shafran, Izhak and Narasimhan, Karthik and Cao, Yuan},
  booktitle = {International Conference on Learning Representations},
  year = {2023},
  url = {https://openreview.net/forum?id=WE_vluYUL-X}
}

@misc{shao2024deepseekmath,
  title = {{DeepSeekMath}: Pushing the Limits of Mathematical Reasoning in Open Language Models},
  author = {Shao, Zhihong and Wang, Peiyi and Zhu, Qihao and Xu, Runxin and Song, Junxiao and Bi, Xiao and Zhang, Haowei and Zhang, Mingchuan and Li, Y. K. and Wu, Y. and Guo, Daya},
  year = {2024},
  eprint = {2402.03300},
  archivePrefix = {arXiv},
  primaryClass = {cs.CL},
  url = {https://arxiv.org/abs/2402.03300}
}

@inproceedings{cao2025icr,
  title = {{ICR}: Iterative Clarification and Rewriting for Conversational Search},
  author = {Cao, Zhiyu and Li, Peifeng and Zhu, Qiaoming},
  booktitle = {Proceedings of the 2025 Conference on Empirical Methods in Natural Language Processing},
  pages = {9810--9824},
  year = {2025},
  publisher = {Association for Computational Linguistics},
  url = {https://aclanthology.org/2025.emnlp-main.496/},
  doi = {10.18653/v1/2025.emnlp-main.496}
}

@inproceedings{qian2024tell,
  title = {Tell Me More! Towards Implicit User Intention Understanding of Language Model Driven Agents},
  author = {Qian, Cheng and He, Bingxiang and Zhuang, Zhong and Deng, Jia and Qin, Yujia and Cong, Xin and Zhang, Zhong and Zhou, Jie and Lin, Yankai and Liu, Zhiyuan and Sun, Maosong},
  booktitle = {Proceedings of the 62nd Annual Meeting of the Association for Computational Linguistics (Volume 1: Long Papers)},
  year = {2024},
  pages = {1088--1113},
  doi = {10.18653/v1/2024.acl-long.61}
}

@inproceedings{zhang2024askbeforeplan,
    title = "Ask-before-Plan: Proactive Language Agents for Real-World Planning",
    author = "Zhang, Xuan  and
      Deng, Yang  and
      Ren, Zifeng  and
      Ng, See-Kiong  and
      Chua, Tat-Seng",
    editor = "Al-Onaizan, Yaser  and
      Bansal, Mohit  and
      Chen, Yun-Nung",
    booktitle = "Findings of the Association for Computational Linguistics: EMNLP 2024",
    month = nov,
    year = "2024",
    address = "Miami, Florida, USA",
    publisher = "Association for Computational Linguistics",
    url = "https://aclanthology.org/2024.findings-emnlp.636/",
    doi = "10.18653/v1/2024.findings-emnlp.636",
    pages = "10836--10863"
}

@misc{yang2025qwen3,
      title={Qwen3 Technical Report}, 
      author={An Yang and Anfeng Li and Baosong Yang and Beichen Zhang and Binyuan Hui and Bo Zheng and Bowen Yu and Chang Gao and Chengen Huang and Chenxu Lv and Chujie Zheng and Dayiheng Liu and Fan Zhou and Fei Huang and Feng Hu and Hao Ge and Haoran Wei and Huan Lin and Jialong Tang and Jian Yang and Jianhong Tu and Jianwei Zhang and Jianxin Yang and Jiaxi Yang and Jing Zhou and Jingren Zhou and Junyang Lin and Kai Dang and Keqin Bao and Kexin Yang and Le Yu and Lianghao Deng and Mei Li and Mingfeng Xue and Mingze Li and Pei Zhang and Peng Wang and Qin Zhu and Rui Men and Ruize Gao and Shixuan Liu and Shuang Luo and Tianhao Li and Tianyi Tang and Wenbiao Yin and Xingzhang Ren and Xinyu Wang and Xinyu Zhang and Xuancheng Ren and Yang Fan and Yang Su and Yichang Zhang and Yinger Zhang and Yu Wan and Yuqiong Liu and Zekun Wang and Zeyu Cui and Zhenru Zhang and Zhipeng Zhou and Zihan Qiu},
      year={2025},
      eprint={2505.09388},
      archivePrefix={arXiv},
      primaryClass={cs.CL},
      url={https://arxiv.org/abs/2505.09388}, 
}

@inproceedings{hu2022lora,
  title={{LoRA}: Low-Rank Adaptation of Large Language Models},
  author={Hu, Edward J. and Shen, Yelong and Wallis, Phillip and Allen-Zhu, Zeyuan and Li, Yuanzhi and Wang, Shean and Wang, Lu and Chen, Weizhu},
  booktitle={International Conference on Learning Representations},
  year={2022},
  url={https://openreview.net/forum?id=nZeVKeeFYf9}
}

@inproceedings{song2020mpnet,
  title={MPNet: Masked and Permuted Pre-training for Language Understanding},
  author={Song, Kaitao and Tan, Xu and Qin, Tao and Lu, Jianfeng and Liu, Tie-Yan},
  booktitle={Advances in Neural Information Processing Systems},
  volume={33},
  year={2020},
  url={https://proceedings.neurips.cc/paper/2020/hash/c3a690be93aa602ee2dc0ccab5b7b67e-Abstract.html}
}

@inproceedings{reimers2019sentencebert,
  title={Sentence-{BERT}: Sentence Embeddings using Siamese {BERT}-Networks},
  author={Reimers, Nils and Gurevych, Iryna},
  booktitle={Proceedings of the 2019 Conference on Empirical Methods in Natural Language Processing and the 9th International Joint Conference on Natural Language Processing (EMNLP-IJCNLP)},
  pages={3982--3992},
  year={2019},
  publisher={Association for Computational Linguistics},
  url={https://aclanthology.org/D19-1410/},
  doi={10.18653/v1/D19-1410}
}

@article{douze2024faiss,
  title={The Faiss Library},
  author={Douze, Matthijs and Guzhva, Alexandr and Deng, Chengqi and Johnson, Jeff and Szilvasy, Gergely and Mazar{\'e}, Pierre-Emmanuel and Lomeli, Maria and Hosseini, Lucas and J{\'e}gou, Herv{\'e}},
  journal={arXiv preprint arXiv:2401.08281},
  year={2024}
}

@misc{openai2026gpt54,
  title={GPT-5.4 Model},
  author={{OpenAI}},
  howpublished={\url{https://developers.openai.com/api/docs/models/gpt-5.4}},
  year={2026}
}

@misc{openai2026o4minideepresearch,
  title={o4-mini-deep-research Model},
  author={{OpenAI}},
  howpublished={\url{https://developers.openai.com/api/docs/models/o4-mini-deep-research}},
  year={2026}
}

@misc{serpapi2026google,
  title={Google Search Engine Results API},
  author={{SerpAPI}},
  howpublished={\url{https://serpapi.com/search-api}},
  year={2026}
}

@misc{jinaai2026reader,
  title={Reader API},
  author={{Jina AI}},
  howpublished={\url{https://jina.ai/en-US/reader/}},
  year={2026}
}

@misc{zhu2025oagentsempiricalstudybuilding,
  title={OAgents: An Empirical Study of Building Effective Agents},
  author={Zhu, He and Qin, Tianrui and Zhu, King and Huang, Heyuan and Guan, Yeyi and Xia, Jinxiang and Yao, Yi and Li, Hanhao and Wang, Ningning and Liu, Pai and Peng, Tianhao and Gui, Xin and Li, Xiaowan and Liu, Yuhui and Jiang, Yuchen Eleanor and Wang, Jun and Zhang, Changwang and Tang, Xiangru and Zhang, Ge and Yang, Jian and Liu, Minghao and Gao, Xitong and Zhou, Wangchunshu and Liu, Jiaheng},
  year={2025},
  eprint={2506.15741},
  archivePrefix={arXiv},
  primaryClass={cs.AI},
  url={https://arxiv.org/abs/2506.15741}
}

@inproceedings{zamani2020clarifying,
  title     = {Generating Clarifying Questions for Information Retrieval},
  author    = {Zamani, Hamed and Dumais, Susan and Craswell, Nick and Bennett, Paul and Lueck, Gord},
  booktitle = {Proceedings of The Web Conference 2020},
  pages     = {418--428},
  year      = {2020},
  publisher = {ACM},
  doi       = {10.1145/3366423.3380126},
  url       = {https://doi.org/10.1145/3366423.3380126}
}

\clearpage
\appendix
\section{Implementation Details}

Code is available at {\codeurl}.

\subsection{Data, Targets, and Trajectories}
Candidate elicitation targets are extracted from the initial user query and benchmark-provided personalization rubrics using GPT-5.4~\citep{openai2026gpt54}.
The extracted targets are validated and organized into intent elicitation graphs that encode their dependencies, so a target is considered only after the targets it depends on have been resolved.
We generate scenario trajectories by varying optional-target inclusion, memory availability, late activation, and stopping conditions.
For each trajectory step, the current evidence state and interaction history are used as input, while the next policy action and its output serve as the SFT target.
Table~\ref{tbl:appendix-data-statistics} summarizes the dataset, target, and training-data statistics.

\begin{table}[h]
\centering
\begin{tabular*}{\linewidth}{@{}l@{\extracolsep{\fill}}rrr@{}}
\toprule
\multicolumn{4}{l}{\textbf{Dataset coverage}} \\
\midrule
Users & \multicolumn{3}{r}{25} \\
Tasks & \multicolumn{3}{r}{30} \\
Instances & \multicolumn{3}{r}{150} \\
\midrule
\rowcolor{gray!15}
\multicolumn{4}{l}{\textbf{Target statistics by split}} \\
 & \textbf{Train} & \textbf{Eval.} & \textbf{All} \\
Instances & 105 & 45 & 150 \\
Elicitation targets & 916 & 418 & 1,334 \\
Elicitation targets / instance & 8.72 & 9.29 & 8.89 \\
Essential  targets & 387 & 172 & 559 \\
Essential targets / instance & 3.69 & 3.82 & 3.73 \\
\midrule
\rowcolor{gray!15}
\multicolumn{4}{l}{\textbf{Training trajectory construction}} \\
Scenario trajectories & \multicolumn{3}{r}{31,081} \\
Step-level SFT examples & \multicolumn{3}{r}{184,324} \\
\bottomrule
\end{tabular*}
\caption{
Dataset, target, and training-data statistics.
Each trajectory yields multiple step-level SFT examples.
}
\label{tbl:appendix-data-statistics}
\end{table}

\subsection{User State Construction}
We construct separate user states for SFT and GRPO~\citep{shao2024deepseekmath} because the two training stages use memory for different purposes.
The offline SFT state is constructed to reproduce the evidence conditions specified by each training trajectory, including whether a target can be resolved from memory or must be acquired through dialogue.
In contrast, the online GRPO state stores only information judged useful beyond the current interaction, approximating the user information that \proposed can accumulate and later retrieve during inference.
Both states distinguish slot-level episodic memories, which are accessed through \textsc{Retrieve}, from compact profile memories supplied as persistent user context.

\paragraph{Offline SFT state.}
Before rendering SFT trajectories, each validated elicitation target is canonicalized into a slot name, purpose, and value.
The resulting memories are separated by user and indexed with FAISS~\citep{douze2024faiss}.
During SFT trajectory rendering, memory availability is determined by the predefined trajectory scenario rather than by information accumulated through previous policy interactions.
When a scenario specifies that a target should be available from memory, the renderer includes the memory corresponding to that target.
Any remaining retrieval slots are filled with memories from the same user's other training instances that have the highest retrieval similarity scores.
When the scenario specifies that the target is not available from memory, only memories from the same user's other instances are returned.
The retriever examines up to 15 candidates and returns at most three memories.

Offline profile memories are constructed separately from episodic memories.
For each user and training instance, GPT-5.4 summarizes reusable information from that user's other instances while excluding all target values belonging to the current instance.
The resulting one-sentence profile allows SFT trajectories to include persistent user context without directly revealing the current instance's target values.

\paragraph{Online GRPO state.}
GRPO begins with empty memory and profile states and updates them only after a completed interaction trajectory.
All trajectories in the same rollout group are generated from an identical user-state snapshot.
After reward computation, only the trajectory receiving the largest total reward within the group can modify the shared user state.
The other trajectories contribute to the GRPO policy update but do not add or modify user information.
A Qwen3-32B judge~\citep{yang2025qwen3} stores a covered slot only when it is directly supported by the interaction, specific to the user, likely to be useful in future interactions, and not already represented in the profile.
The same judge optionally updates the profile when the newly observed information indicates a durable user preference, goal, or constraint that is applicable beyond the current task.

For evaluation, the memory and profile states accumulated during GRPO training are frozen and used as the initial test-time user state.
Users observed during training retain their accumulated states, whereas unseen users begin with empty states.
These states are constructed exclusively from training-split interactions, and test interactions do not modify them.

\subsection{User Simulator}
We use Qwen3-32B~\citep{yang2025qwen3} as a controlled user simulator for interactive rollouts.
For each instance, the simulator receives task-relevant facts from the PDR-Bench persona~\citep{pdrbench}, together with the elicitation targets and their question and answer seeds.
This simulator-only information is not exposed to the agent, which observes only its current profile, retrieved memories, and dialogue history.
When the agent asks a clarification question, the simulator identifies the most relevant target using its description and question seed, and generates a response grounded in the corresponding answer seed and persona information.
If neither the targets nor the persona support an answer, the simulator returns a brief non-committal response rather than introducing a new user preference.

\begin{table*}[t]
\centering
{\small
\begin{tabular}{p{0.20\textwidth}p{0.36\textwidth}p{0.36\textwidth}}
\toprule
Setting & SFT & GRPO \\
\midrule
Initialization
& \texttt{Qwen/Qwen3-8B}
& Final SFT checkpoint \\
Learning rate
& $2\times10^{-4}$
& $1\times10^{-6}$ \\
Optimization
& AdamW, weight decay 0.0
& AdamW, weight decay 0.01 \\
Training length
& 2 epochs
& 2 passes over 105 training instances (210 updates) \\
Batching
& 64 examples per update
& 4 trajectories per training instance \\
LoRA
& Rank 8, alpha 16, dropout 0.05; attention and MLP projections
& Same as SFT \\
Precision
& bf16 with a 4-bit NF4-quantized base model
& Same as SFT \\
Random seed
& 42
& 42 \\
\bottomrule
\end{tabular}
}
\caption{Final training settings for SFT and GRPO.}
\label{tbl:training-config}
\end{table*}

\subsection{Training Configuration}
The policy is initialized from Qwen3-8B~\citep{yang2025qwen3} and trained with LoRA adapters~\citep{hu2022lora}.
Qwen3-32B supports user simulation and online memory and profile updates but is not fine-tuned.
GPT-5.4 is used for target and graph construction, offline profile summarization, evaluation-time factor extraction, and downstream P/Q scoring.
It is not used as the trainable policy or user simulator.
Table~\ref{tbl:training-config} summarizes the final SFT and GRPO settings.

SFT provides step-level demonstrations under controlled evidence states, teaching the policy when to retrieve, ask, or stop.
GRPO then samples groups of interaction trajectories and optimizes the policy using their relative rewards while accumulating reusable user information through completed rollouts.
The reward combines semantic target coverage with false-positive, action-cost, repetition, multi-question, invalid-output, and non-stop penalties.
All trainable variants use a single training run with seed 42.

\paragraph{GRPO-Specific Settings.}
GRPO uses a group size of four, a clipping parameter of 0.2, and no KL penalty.
Rollouts use temperature 0.7 and top-$p$ 0.95.
The reward coefficients are
$\alpha_{\mathrm{align}}=1.0$,
$\beta_{\mathrm{fp}}=0.3$,
$\lambda_{\mathrm{ask}}=0.04$,
$\lambda_{\mathrm{ret}}=0.01$,
$\lambda_{\mathrm{rep}}=0.08$, and
$\lambda_{\mathrm{multi}}=0.08$.
Invalid-output and non-stop penalties are 1.0 and 0.2, respectively.
Repeated-question detection uses a cosine-similarity threshold of 0.86.

\paragraph{Compute Environment.}
SFT used four NVIDIA A100-SXM4-80GB GPUs, whereas GRPO used one
NVIDIA A100-SXM4-80GB GPU.
Experiments ran on Linux 5.14 with Python 3.10.19, PyTorch 2.7.0
with CUDA 12.6, Transformers 5.8.1, PEFT 0.18.1, TRL 1.4.0,
bitsandbytes 0.49.2, FAISS 1.12.0, and sentence-transformers 5.2.3.

\section{Evaluation Details}

\subsection{Rollout and Factor Extraction}
At test time, \proposed uses temperature 0.7, top-$p$ 1.0, and at most 1024 new agent tokens.
We use GPT-5.4~\citep{openai2026gpt54} through the OpenAI API with a fixed prompt to identify the user-specific framing factors expressed in each final refined query and return them in a structured format.
These requests use the model identifier \texttt{gpt-5.4-}\allowbreak\texttt{2026-03-05}.
Each final refined query is submitted once, and every response is successfully parsed in the required format.
For coverage, each reference target is compared with its most similar extracted factor.
For precision, each extracted factor is compared with its most similar reference target.
The two matching directions are computed independently.
Reported metrics are computed by pooling the corresponding matched and total counts or weights across instances within each evaluation group, rather than averaging instance-level scores.
The same prompt and decoding settings are used for every call that identifies user-specific framing factors.

\begin{table*}[t]
\centering
{\small
\setlength{\tabcolsep}{4pt}
\begin{tabularx}{\textwidth}{
@{}
>{\raggedright\arraybackslash}p{0.17\textwidth}
>{\raggedright\arraybackslash}p{0.21\textwidth}
>{\raggedright\arraybackslash}p{0.17\textwidth}
>{\raggedright\arraybackslash}p{0.23\textwidth}
>{\raggedright\arraybackslash}X
@{}}
\toprule
\textbf{Method}
& \textbf{Policy / checkpoint}
& \textbf{Action interface}
& \textbf{Test-time user state}
& \textbf{Adaptation} \\
\midrule

Direct Rewrite
& Qwen3-8B
& \textsc{Stop}
& Frozen profile
& Prompting only \\

Memory-RAG Agent
& Qwen3-8B
& \textsc{Retrieve}/\textsc{Stop}
& Frozen profile and episodic memory
& Prompting only \\

Clarification Agent
& Qwen3-8B
& \textsc{Ask}/\textsc{Stop}
& Frozen profile
& Prompting only \\

Memory+Clarification Agent
& Qwen3-8B
& \textsc{Retrieve}/\textsc{Ask}/\textsc{Stop}
& Frozen profile and episodic memory
& Prompting only \\

\addlinespace

Mistral-Interact
& Released Mistral-Interact checkpoint
& \textsc{Ask}/\textsc{Stop}
& None
& Released checkpoint \\

CEP-Clarify+Rewrite
& Llama-3-8B-Instruct with the released CEP checkpoint
& \textsc{Ask}/\textsc{Stop}
& None
& Released checkpoint \\

IntentRL-style Clarifier
& Qwen3-8B + LoRA
& \textsc{Ask}/\textsc{Stop}
& Frozen profile
& IEG-scaffolded SFT + GRPO \\

\proposed
& Qwen3-8B + LoRA
& \textsc{Retrieve}/\textsc{Ask}/\textsc{Stop}
& Frozen profile and episodic memory
& IEG-scaffolded SFT + GRPO \\

\bottomrule
\end{tabularx}
}
\caption{
Implementation comparison of evaluated methods.
For Mistral-Interact and CEP-Clarify+Rewrite, their native interaction outputs are mapped to the common \textsc{Ask}/\textsc{Stop} interface.
}
\label{tbl:baseline-implementation}
\end{table*}

\subsection{Baseline Implementations}

\paragraph{Prompting and Released Baselines.}
The Qwen prompting baselines use the same Qwen3-8B~\citep{yang2025qwen3} backbone and differ only in their system prompts and allowed action menus.
Interactive methods use the same Qwen3-32B user simulator with temperature 0.7, top-$p$ 1.0, and at most 768 new tokens.
Direct Rewrite and Memory-RAG Agent do not call the simulator because they never ask the user.
The Memory+Clarification Agent is given the same frozen online profile, FAISS memory index, and \textsc{Retrieve} action as the Memory-RAG Agent.
We use the released Mistral-Interact checkpoint~\citep{qian2024tell} and the released CEP checkpoint~\citep{zhang2024askbeforeplan}.
Table~\ref{tbl:baseline-implementation} summarizes the policy, action interface, test-time user state, and adaptation strategy of each evaluated method.

\paragraph{IntentRL-Style Baseline.}
We adapt IntentRL~\citep{intentrl} to our setting rather than directly reproducing it.
The original CDAG specifies which clarification question to ask but does not provide supervision for when to stop.
We therefore augment the graph-based training trajectories with explicit stopping decisions, yielding an \textsc{Ask}/\textsc{Stop} policy.
To control for differences in the backbone and optimization procedure, the baseline uses Qwen3-8B with LoRA, a frozen user profile, an IEG-scaffolded SFT warm-up, and GRPO optimization.
During evaluation, it observes the frozen user profile and dialogue history but receives neither the validated elicitation targets nor the offline IEG.

\begin{table*}[t]
\centering
\begin{tabularx}{\linewidth}{@{}ll *{8}{>{\centering\arraybackslash}X}@{}}
    \toprule
    \multirow{2}{*}{\makecell{\textbf{Downstream}\\\textbf{DRA}}}
    & \multirow{2}{*}{\textbf{Method}}
    & \multicolumn{2}{c}{\textbf{Overall}}
    & \multicolumn{2}{c}{\textbf{Unseen User}}
    & \multicolumn{2}{c}{\textbf{Unseen Task}}
    & \multicolumn{2}{c}{\textbf{Both Unseen}} \\
    \cmidrule(lr){3-4} \cmidrule(lr){5-6} \cmidrule(lr){7-8} \cmidrule(lr){9-10}
    & & P & Q & P & Q & P & Q & P & Q \\
    \midrule
    \multirow{6}{*}{\makecell[l]{OpenAI\\Deep Research}}
    & Original Query
    & 3.70 & 5.43
    & 3.74 & 5.44
    & 3.60 & 5.34
    & 3.66 & 5.40 \\

    & Memory+Clarification
    & 3.81 & 5.28
    & 3.80 & 5.23
    & 3.62 & 5.12
    & 3.76 & 5.19 \\

    & Mistral-Interact
    & 3.28 & 4.71
    & 3.15 & 4.63
    & 2.95 & 4.39
    & 2.98 & 4.35 \\

    & CEP-Clarify+Rewrite
    & \underline{4.25} & 5.40
    & \underline{4.09} & 5.34
    & 3.98 & 5.26
    & \underline{4.10} & \underline{5.43} \\

    & IntentRL-style Clarifier
    & 4.22 & \textbf{5.69}
    & 4.00 & \underline{5.48}
    & \underline{4.06} & \textbf{5.63}
    & 3.91 & \textbf{5.57} \\

    \rowcolor{gray!12}
    \cellcolor{white} & \textbf{\proposed}
    & \textbf{4.35} & \underline{5.56}
    & \textbf{4.40} & \textbf{5.59}
    & \textbf{4.10} & \underline{5.38}
    & \textbf{4.12} & 5.38 \\

    \midrule
    \multirow{6}{*}{OAgents}
    & Original Query
    & 5.05 & \underline{6.73}
    & 4.71 & 6.41
    & 5.02 & \underline{6.63}
    & 4.66 & 6.22 \\

    & Memory+Clarification
    & 5.12 & 6.37
    & 5.05 & 6.38
    & 4.72 & 6.13
    & 4.68 & 6.19 \\

    & Mistral-Interact
    & 4.94 & 6.05
    & 4.74 & 6.03
    & 4.63 & 5.80
    & 4.62 & 5.78 \\

    & CEP-Clarify+Rewrite
    & \underline{5.53} & 6.37
    & \textbf{5.73} & \textbf{6.89}
    & \underline{5.31} & 6.22
    & \textbf{5.68} & \underline{6.65} \\

    & IntentRL-style Clarifier
    & 4.93 & 5.92
    & 4.56 & 5.62
    & 4.49 & 5.42
    & 4.05 & 4.86 \\

    \rowcolor{gray!12}
    \cellcolor{white} & \textbf{\proposed}
    & \textbf{5.69} & \textbf{6.75}
    & \underline{5.65} & \underline{6.86}
    & \textbf{5.51} & \textbf{6.66}
    & \underline{5.63} & \textbf{6.79} \\

    \bottomrule
\end{tabularx}
\caption{
Downstream P/Q scores overall and across unseen-user and unseen-task settings.
Bold and underline indicate the best and second-best results within each DRA.
}
\label{tbl:downstream-pq-app}
\end{table*}

\subsection{Downstream Report Setup}
We evaluate downstream report utility using the PDR-Bench P/Q evaluation protocol~\citep{pdrbench}.
For each method on each evaluation instance, each downstream deep research agent (DRA) receives the final refined query and generates one report.
OpenAI Deep Research uses the Responses API model identifier \texttt{o4-mini-deep-research-}\allowbreak\texttt{2025-06-26}~\citep{openai2026o4minideepresearch}, with at most 15 web-search tool calls.
OAgents~\citep{zhu2025oagentsempiricalstudybuilding} uses \texttt{gpt-5-mini} with its crawler web stack, SerpAPI-backed web search~\citep{serpapi2026google}, Jina read mode~\citep{jinaai2026reader}, at most 8 agent steps, and at most 8 search steps.
Following IntentRL~\citep{intentrl}, we report personalization alignment (P) and content quality (Q), the PDR-Bench dimensions most directly related to the effects of pre-research clarification.
We omit factual reliability (R), which evaluates claim support and citation quality and depends largely on the fixed DRA's downstream search and verification behavior.
The PDR-Bench evaluator uses GPT-5.4, with completed response artifacts reporting the model identifier \texttt{gpt-5.4-2026-03-05}.
For each generated report and each metric, we run three independent scoring rounds and report their average.
Because the resulting scores reflect both the refined query and DRA-specific planning, retrieval, and synthesis, we compare methods within each DRA rather than comparing absolute scores across DRAs.

\section{Additional Results}
\begin{table*}[t]
\centering
\begin{tabular*}{\linewidth}{@{}l@{\extracolsep{\fill}}lcccccc@{}}
\toprule
\textbf{Metric} & \textbf{Method} & 0.45 & 0.50 & 0.55 & 0.60 & 0.65 & 0.70 \\
\midrule
WCov & Direct Rewrite & 0.7498 & 0.5473 & 0.3305 & 0.1673 & 0.0719 & 0.0285 \\
 & Memory+Clarification Agent & 0.7130 & 0.5445 & 0.3822 & 0.1965 & 0.0923 & 0.0273 \\
 & Mistral-Interact & 0.7429 & 0.5746 & 0.3811 & 0.1939 & 0.0567 & 0.0161 \\
 & CEP-Clarify+Rewrite & 0.8134 & 0.6416 & 0.4060 & 0.2410 & 0.1324 & 0.0514 \\
 & IntentRL-style Clarifier & 0.7375 & 0.5665 & 0.3760 & 0.2119 & 0.0875 & 0.0275 \\
 & \proposed & 0.7787 & 0.6235 & 0.4253 & 0.2345 & 0.1031 & 0.0390 \\
\midrule
E-WCov & Direct Rewrite & 0.7750 & 0.5873 & 0.3697 & 0.1839 & 0.0795 & 0.0281 \\
 & Memory+Clarification Agent & 0.7619 & 0.5687 & 0.4364 & 0.2349 & 0.1047 & 0.0359 \\
 & Mistral-Interact & 0.7836 & 0.6617 & 0.4223 & 0.2186 & 0.0475 & 0.0023 \\
 & CEP-Clarify+Rewrite & 0.8715 & 0.7114 & 0.4497 & 0.2718 & 0.1474 & 0.0514 \\
 & IntentRL-style Clarifier & 0.7806 & 0.6215 & 0.4103 & 0.2381 & 0.0862 & 0.0215 \\
 & \proposed & 0.8135 & 0.6732 & 0.4639 & 0.2704 & 0.0861 & 0.0400 \\
\midrule
F1 & Direct Rewrite & 0.7705 & 0.5547 & 0.3322 & 0.1699 & 0.0719 & 0.0287 \\
 & Memory+Clarification Agent & 0.7475 & 0.5661 & 0.3820 & 0.1978 & 0.0906 & 0.0262 \\
 & Mistral-Interact & 0.7437 & 0.5604 & 0.3841 & 0.2033 & 0.0691 & 0.0216 \\
 & CEP-Clarify+Rewrite & 0.7933 & 0.6096 & 0.3823 & 0.2331 & 0.1247 & 0.0489 \\
 & IntentRL-style Clarifier & 0.7616 & 0.5893 & 0.3775 & 0.2052 & 0.0907 & 0.0274 \\
 & \proposed & 0.7693 & 0.6082 & 0.4082 & 0.2092 & 0.0949 & 0.0347 \\
\bottomrule
\end{tabular*}
\caption{Threshold sensitivity of absolute scores across cosine similarity thresholds.}
\label{tbl:threshold_sensitivity_scores}
\end{table*}

\subsection{Downstream Performance on Unseen Settings}
Table~\ref{tbl:downstream-pq-app} reports downstream personalization and quality overall and for the Unseen User, Unseen Task, and Both Unseen settings.
The Unseen User results combine User-only unseen and Both unseen instances, while the Unseen Task results combine Task-only unseen and Both unseen instances.

With OpenAI Deep Research, \proposed achieves the highest personalization score overall and in all three unseen settings.
It also preserves competitive report quality, obtaining the highest Q score for unseen users while remaining close to the strongest alternatives overall and on unseen tasks.
With OAgents, \proposed achieves the highest overall personalization and quality scores.
It also performs best on both metrics for unseen tasks and achieves the highest quality score when both the user and task are unseen.
CEP-Clarify+Rewrite is slightly stronger in personalization for unseen users and both-unseen instances.
These exceptions show that the relative advantage of each refinement method can vary with the downstream DRA and the type of distribution shift.
Overall, \proposed maintains strong personalization under user and task shift, while report quality remains competitive across downstream DRAs and unseen settings.

\subsection{Threshold Sensitivity}

The main evaluation uses a cosine-similarity threshold of 0.55, representing an intermediate setting between the more permissive and stricter thresholds examined.
To examine how sensitive the evaluation results are to this threshold, Table~\ref{tbl:threshold_sensitivity_scores} reports absolute WCov, E-WCov, and F1 scores for thresholds from 0.45 to 0.70.
As the threshold becomes stricter, absolute scores generally decrease because fewer semantic matches are accepted.
The relative performance of the methods varies across metrics and thresholds, and we therefore do not claim uniform superiority over the full range.

\section{Artifacts, Licenses, and Data Handling}
We evaluate on the processed English subset of PDR-Bench~\citep{pdrbench},
whose release and Hugging Face dataset are licensed under Apache-2.0.
Qwen3-8B, Qwen3-32B, and
\texttt{sentence-transformers/all-mpnet-base-v2}
\citep{reimers2019sentencebert,song2020mpnet} are also licensed under
Apache-2.0, while FAISS~\citep{douze2024faiss} uses the MIT License.
We use these artifacts only for research, do not deploy the resulting memory
or profile representations, collect no additional human-subject data, and
recruit no external annotators.
Memory and profiles contain only request-relevant information grounded in
available user evidence.
We neither use them to infer protected attributes nor release raw records
containing direct identifiers or sensitive details.




\end{document}